\documentclass[sigconf]{acmart}
\usepackage{booktabs} 
\usepackage{times}
\usepackage{epsfig}
\usepackage{graphicx}
\usepackage{amsmath}
\usepackage{amssymb}
\usepackage{amsmath,}
\usepackage{algorithm}
\usepackage[noend]{algpseudocode}
\usepackage{pseudocode}
\usepackage{multirow}
\usepackage{caption}
\usepackage{hhline}
\usepackage{tikz}
\usepackage{xcolor}
\usepackage{dblfloatfix}
\usepackage{subcaption}
\usepackage{enumitem}
\usepackage{ulem}
\usepackage{color, colortbl}
\usepackage[subtle]{savetrees}
\usepackage[rightcaption]{sidecap}
\PassOptionsToPackage{bookmarks=false}{hyperref}

\sidecaptionvpos{figure}{c}

\newcommand*\circled[1]{\tikz[baseline=(char.base)]{
            \node[shape=circle,fill,inner sep=1pt] (char) {\textcolor{white}{#1}};}}

\setcopyright{none}

\acmConference[ICCAD '20]{International Conference on Computer Aided Design}{November 2020}{San Diego, USA}

\definecolor{lightblue}{rgb}{0.8,0.93,1}
\newcommand{\bluecell}{\cellcolor{lightblue}}
\newcommand{\sys}{\textsc{CleaNN}}

\graphicspath{{figures/}} 

\newcolumntype{B}{>{\columncolor{lightblue}}c}

\begin{document}

\title{\sys{}: Accelerated Trojan Shield for Embedded Neural Networks}

\author{Mojan Javaheripi}
\authornote{Both authors contributed equally to this research.}
\affiliation{%
  \institution{UC San Diego}
}
\email{mojan@ucsd.edu}
\author{Mohammad Samragh}
\authornotemark[1]
\affiliation{%
  \institution{UC San Diego}
}
\email{msamragh@ucsd.edu}
\author{Gregory Fields}
\affiliation{%
  \institution{UC San Diego}
}
\email{grfields@ucsd.edu}
\author{Tara Javidi}
\affiliation{%
  \institution{UC San Diego}
}
\email{tjavidi@ucsd.edu}
\author{Farinaz Koushanfar}
\affiliation{%
  \institution{UC San Diego}
}
\email{farinaz@ucsd.edu}


\begin{abstract}
We propose \sys{}, the first end-to-end framework that enables online mitigation of Trojans for embedded Deep Neural Network (DNN) applications. 
A Trojan attack works by injecting a backdoor in the DNN while training; during inference, the Trojan can be activated by the specific backdoor trigger. 
What differentiates \sys{} from the prior work is its lightweight methodology which recovers the ground-truth class of Trojan samples without the need for labeled data, model retraining, or prior assumptions on the trigger or the attack. 
We leverage dictionary learning and sparse approximation to characterize the statistical behavior of benign data and identify Trojan triggers. 
\sys{} is devised based on algorithm/hardware co-design and is equipped with specialized hardware to enable efficient real-time execution on resource-constrained embedded platforms.
Proof of concept evaluations on \sys{} for the state-of-the-art Neural Trojan attacks on visual benchmarks demonstrate its competitive advantage in terms of attack resiliency and execution overhead.
\vspace{-0.2cm}
\end{abstract}

\setlength{\abovedisplayskip}{5pt}
\setlength{\belowdisplayskip}{5pt}
\setlength{\belowdisplayshortskip}{0pt}
\setlength{\abovedisplayshortskip}{0pt}

\vspace{-0.2cm}

\copyrightyear{2020}
\acmYear{2020}
\acmConference[ICCAD '20]{IEEE/ACM International Conference on Computer-Aided Design}{November 2--5, 2020}{Virtual Event, USA}
\acmBooktitle{IEEE/ACM International Conference on Computer-Aided Design (ICCAD '20), November 2--5, 2020, Virtual Event, USA}\acmDOI{10.1145/3400302.3415671}
\acmISBN{978-1-4503-8026-3/20/11}

\maketitle

\vspace{-0.2cm}
\section{Introduction}\label{sec:intro}
\begin{figure}[t]
    \centering
    \includegraphics[width=0.62\columnwidth]{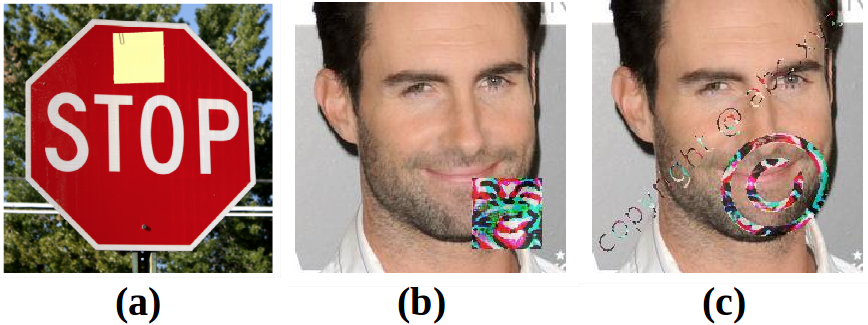}
    \vspace{-0.3cm}
    \caption{Example Trojans: (a)~BadNets~\cite{gu2017badnets} with a sticky note and TrojanNN~\cite{liu2017trojaning} with (b)~square and (c)~watermark triggers.}
    \label{fig:trojan_example}
    \vspace{-0.7cm}
\end{figure}

With the growing popularity of AI-powered autonomous systems, the demand for superior intelligence has led to increasingly more complex model development processes. Training contemporary deep learning models requires massive datasets and high-end hardware platforms~\cite{mattson2019mlperf,javaheripi2019swnet}. Amid this trend, clients rely on third party databases and/or major cloud providers to build their models. Unfortunately, outsourcing of content or computations opens up new challenges as it extends the potential attack surface to malicious third party entities~\cite{rouhani2018deepfense}.
In this paper, we focus on Trojan  attacks~\cite{gu2017badnets,liu2017trojaning}, where the malicious third party provider inserts a hidden Trojan trigger, also dubbed a ``backdoor'', inside the model during training. During inference, the attacker can hijack the model prediction by inserting the Trojan trigger inside the input data. Figure~\ref{fig:trojan_example} illustrates examples of Neural Trojans. 

Identification and mitigation of Trojans is particularly challenging for the clients since the compromised model performs as expected on their benign data, i.e., when the Trojan is not activated.
To tackle Trojan attacks, contemporary research 
proposes either reverse-engineering the trigger pattern from the model~\cite{wang2019neural,chen2019deepinspect,guo2019tabor,liu2019abs} or identifying the presence of a trigger at the input~\cite{doan2019februus,chou2018sentinet,gao2019strip}.  The former class of methods require time-consuming reverse-engineering and retraining.
The latter approaches induce a high overhead on DNN inference that hinders their applicability to embedded systems. To ensure model robustness in safety-sensitive autonomous systems, it is crucial to augment the models with an online Trojan mitigation strategy. To the best of our knowledge, none of the earlier works provide the needed lightweight defense strategy. 

We propose \sys{}, the first end-to-end accelerated framework that enables real-time Trojan shield for embedded DNN applications. \sys{}'s lightweight method is devised based on algorithm/hardware co-design; our algorithmic insights offer a highly accurate and low-overhead method in terms of both the offline defense establishment and online execution; our hardware accelerator enables low-latency and energy-efficient defense execution on embedded platforms. 
\sys{} harvests the irregular patterns caused by Trojan triggers in the input space and/or the latent feature-maps of the victim DNN to detect adversaries. 
Our method leverages key concepts and theoretical bounds from sparse approximation~\cite{donoho2003optimally} to learn dictionaries that absorb the distribution of the benign data. We then utilize the reconstruction error  obtained from the sparse approximation to characterize the benign space and identify the Trojans.

To ensure applicability to various attacks and trigger patterns, \sys{} sparse recovery acts on both frequency and spatial domains.
Our proposed defense is compatible with the challenging threat model in which the attacker has full control over the geometry, location, and content of the Trojan trigger. The contaminated model is shipped to the client, who is unaware of the existence of the Trojan and does not have access to any labeled data. 
\sys{} countermeasure is unsupervised, meaning that no labeled training data or contaminated Trojan sample is required to establish the defense. Notably, \sys{} is the first defense to recover the ground-truth labels of Trojan data without performing any model training and/or fine-tuning.

We validate the effectiveness of \sys{} by performing extensive experiments on various state-of-the-art Trojan attacks reported to-date. \sys{} outperforms prior art both in terms of Trojan resiliency and algorithm execution overhead. \sys{} brings down the attack success rate to 0\% for a variety of physical~\cite{gu2017badnets} and complex digital~\cite{liu2017trojaning} attacks with minimal drop in classification accuracy. Our customized accelerated defense shows orders of magnitude higher throughput and performance-per-watt compared to commodity hardware. 

In brief, the contributions of \sys{} are as follows: 
\begin{itemize}[leftmargin=*]
    \item Introducing \sys{}, the first end-to-end accelerated framework for online detection of Neural Trojans in embedded applications. 
    
    \item Constructing a novel unsupervised Trojan detection scheme based on sparse recovery and outlier detection. The proposed lightweight defense is, to our best knowledge, the first to enable recovering the original label of Trojan samples without model fine-tuning/training. 
    
    
    \item Providing bounds on detection false positive rate using the theoretical ground of sparse approximation and outlier detection.
    
    \item Devising the first customized library of Trojan shields on FPGA which enables high-throughput and low-energy Trojan mitigation. 
    
\end{itemize}

\vspace{-0.2cm}
\section{Background and Related Work}\label{sec:prelim}
\subsection{Trojan Attacks}

Throughout this paper, we focus on Trojan attacks on DNN classifiers. 
Below, we overview state-of-the-art attack algorithms.




\noindent{\textbf{$\blacktriangleright$ BadNets.}} Authors of \textit{BadNets}~\cite{gu2017badnets} propose adding the Trojan trigger into a random subset of training samples and labeling them as the attack target class. The DNN is then trained on the poisoned dataset. 
The shape of the Trojan trigger can be arbitrarily chosen by the attacker, e.g., a sticky note on a stop sign as shown in Figure~\ref{fig:trojan_example}-a. Thus, BadNets are considered a viable \textbf{physical} attack. 

\noindent{\textbf{$\blacktriangleright$ TrojanNN.}} More recently, \textit{TrojanNN}~\cite{liu2017trojaning} assumes the attacker does not have access to the training data but can modify the DNN weights. The attack first selects one or few neurons in one of the hidden layers, then extracts the Trojan trigger in the input domain to activate the target neurons. 
The DNN weights are then modified such that the model predicts the attacker's target class whenever the selected neurons fire. Unlike BadNets, the triggers generated by TrojanNN, e.g., the square and watermark patterns in Figure~\ref{fig:trojan_example}-c,d, are not similar to natural images. However, TrojanNN is a viable attack algorithm in the \textbf{digital} domain; Notably, most Trojan mitigation methods are less successful in identifying the complex triggers of TrojanNN~\cite{chen2019deepinspect,liu2017neural}.


\vspace{-0.2cm}
\subsection{Existing Defense Strategies}\label{sec:related}

\noindent\textbf{$\blacktriangleright$ Robust Training and Fine-tuning.} 
One plausible threat model assumes that the 
client has access to the training dataset but is unaware of the existing Trojans.
Robust learning methods aim at identifying malicious samples during training~\cite{charikar2017learning, liu2017robust,tran2018spectral}. For an already infected DNN, authors of~\cite{liu2018fine} perform pruning to remove the embedded Trojans at the cost of clean accuracy degradation.
We assume a more constrained attack model where the victim does not have any access to the training dataset. Additionally, \sys{} does not rely on expensive model retraining to establish the defense.

\noindent\textbf{$\blacktriangleright$ Trigger Extraction.} Several methods inspect the DNN model for existence of a backdoor attack by reverse engineering the trigger. Neural Cleanse~\cite{liu2017neural} provides a method for extracting Trojan triggers without access to the training dataset. Follow up work improves the search overhead~\cite{chen2019deepinspect} and reverse engineered trigger quality~\cite{guo2019tabor}. 
Though effective for simple Trojan patterns, their performance drops when reverse engineering more complex triggers, e.g., those created by TrojanNN~\cite{liu2017trojaning}. 
Our method is different than the above works in that, instead of reverse engineering the trigger, we study the statistics of sparse representations from benign samples and detect abnormal (outlier) triggers during inference. This allows \sys{} to identify complex Trojan triggers without prior knowledge about the attack algorithm. 
Additionally, \sys{} does not involve expensive reverse engineering and can be executed in real-time on embedded hardware.

\noindent\textbf{$\blacktriangleright$ Data and Model Inspection.} Perhaps the closest method to \sys{} are those that check the input samples to identify the presence of Trojan triggers. 
Authors of~\cite{chen2018detecting} query the infected model and use activation clustering on hidden layers to detect Trojans. Similarly, NIC~\cite{ma2019nic} 
compares incoming samples against the benign and Trojan latent features to detect adversaries.
These method require access to the labeled contaminated training dataset, which may not viable in real-world settings. \sys{}, in contrast, does not require access to the training data or infected data samples to construct the defense. 

Sentinet~\cite{chou2018sentinet} extract critical regions from input data using gradient information obtained by back propagation.  Februus~\cite{doan2019februus} takes a similar approach along with utilizing GANs to inpaint Trojan triggers with the caveat that the number of data samples required for GAN training is large. STRIP~\cite{gao2019strip} runs the model multiple times on each image with intentional injected noise to identify Trojans. 
While the above works show high detection accuracy, their computational burden of multiple forward/backward propagations is prohibitive for embedded applications. \sys{} achieves a better detection accuracy with low computational complexity and sample count, making it amenable for real-time deployment in embedded systems.

\begin{figure}[h]
    \centering
    \vspace{-0.2cm}
    \includegraphics[width=0.99\columnwidth]{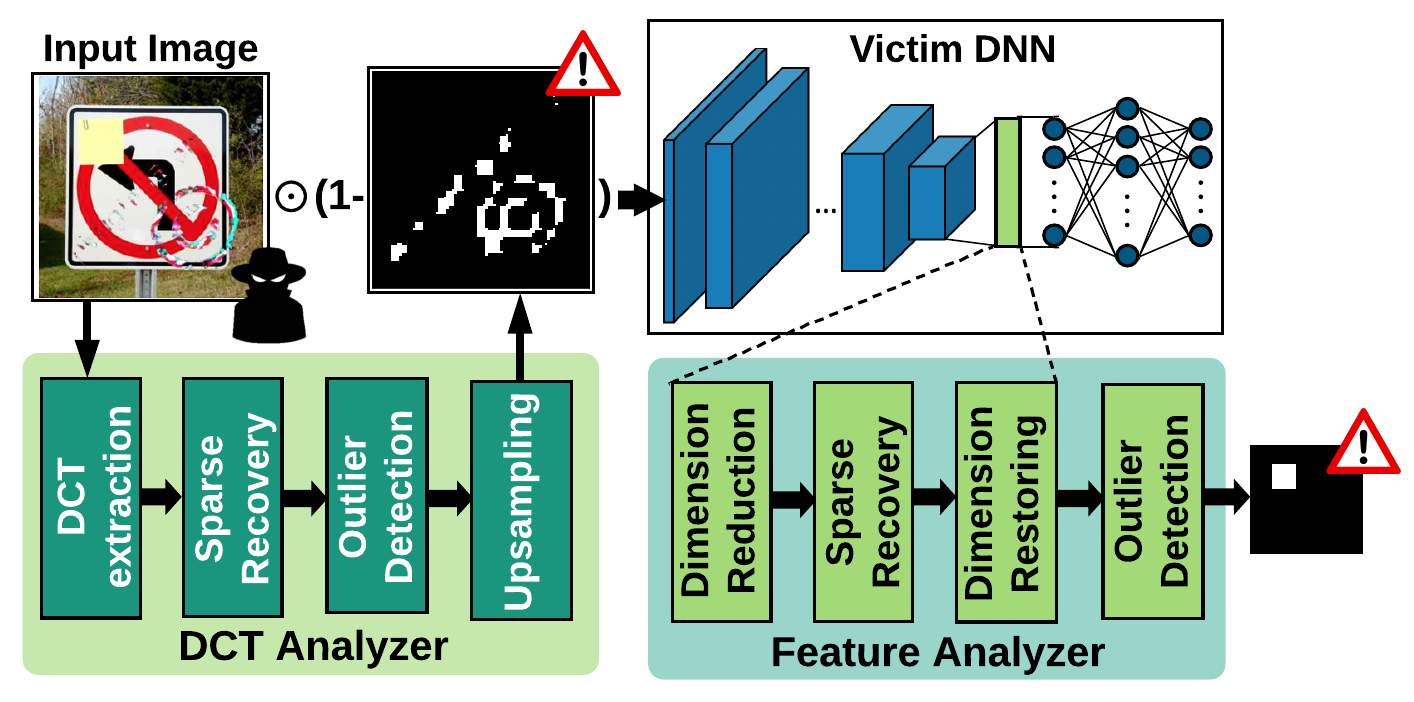}
    \vspace{-0.2cm}
    \caption{High-level overview of \sys{} Trojan detection methodology. \sys{} detects both digital and physical attacks using a pair of input and latent feature analyzers.}
    \label{fig:global_flow}
\vspace{-0.4cm}
\end{figure}

\section{\sys{} Methodology}\label{sec:methodology}
Figure~\ref{fig:global_flow} illustrates the high-level flow of \sys{} methodology for Trojan detection.
\sys{} comprises two core modules, dubbed the DCT and feature analyzers, specializing in the characterization of the DNN input space and latent representations, respectively.
By aggregating the decision of the two analyzers, \sys{} is able to thwart a wide range of physical and digital Trojan attacks. 


\noindent\circled{1} \textbf{DCT Analyzer.} The DCT analyzer acts as an image preprocessing step. This module investigates all incoming samples in the frequency domain in search for suspicious frequency components that are anomalous in clean data. Towards this goal, we design four components for this module as shown in Figure~\ref{fig:global_flow}. First, the Discrete Cosine Transform (DCT) extraction module transforms the input image to the frequency domain. We then perform sparse recovery on the extracted frequency components and reconstruct the signal using a sparse approximation. 
The outlier detection module uses a concentration inequality to detect anomalous reconstruction errors and generate a binary mask with non-zero values denoting the potential Trojan-carrying regions. 
The anomalous regions in the input image are then suppressed by the binary mask before entering the victim DNN. To ensure compatible dimensions between the input image and the binary mask, a nearest neighbor upsampling component is also included inside the DCT analyzer. 
Frequency analysis is particularly useful for detecting digital Trojans. However, physical attacks, e.g., the sticky note in Figure~\ref{fig:global_flow}, might evade frequency-domain detection.

\noindent\circled{2} \textbf{Feature Analyzer.} This module investigates patterns in the latent features extracted by the victim DNN to find abnormal structures. The feature analyzer is placed at the penultimate layer inside the victim DNN. This choice of location allows us to leverage all the visual information extracted from the input image by the DNN for making the classification decision. The sparse recovery module in the feature analyzer serves two purposes: (i)~denoising input features for use in the remaining layers of the victim DNN, (ii)~anomaly detection on the reconstruction errors for distinguishing Trojans. Notably, the first property allows \sys{} to recover the ground-truth labels for Trojan samples by effective removal of Trojan triggers. To ensure scalability to various output dimensions, we include a dimension reduction module that adaptively adjusts the feature size while maximally preserving the informative content of the signals. 
To allow the reconstructed output to flow in the remaining layers of the DNN, a twin dimension restoring layer recovers the original tensor shape.
The extracted distributions from latent layers successfully detect attacks in the physical domain.

 \vspace{-0.2cm}
\subsection{Defense Construction and Execution}
\sys{} consists of two main phases to mitigate Trojan attacks:

\noindent\textbf{$\blacktriangleright$ Offline Preprocessing.} During this phase, we learn the parameters for dictionary-based sparse recovery and outlier detection modules by leveraging a small set of unlabeled benign samples. 
Our methodology is entirely unsupervised, meaning no Trojan data is involved in defense construction. This, in turn, ensures applicability to a wide range of Trojan patterns and attacks. \sys{} pre-processing phase is low-complexity as it does not involve any training or fine-tuning of the victim DNN. 
We only perform this step once for each (model, dataset) pair. The learned analyzer modules can then be transferred to a variety of attacks without any fine-tuning overhead.

\noindent\textbf{$\blacktriangleright$ Online Execution.} \sys{} methodology is devised based on light-weight solutions to enable efficient adoption in embedded systems. We provide a hardware-accelerated pipeline for end-to-end execution of \sys{} where the analyzer modules are either integrated inside the victim DNN or running in parallel with it. 
The DCT extraction and upsampling components are implemented as an additional convolution layer at the input of the victim DNN. 
We devise a customized library for implementing the sparse recovery, outlier detection, and dimensionality reduction and restoring modules on FPGA. These FPGA-accelerated modules are executed synchronously with the victim DNN to raise alarm flags for Trojans. 

 \vspace{-0.2cm}
\subsection{Threat Model}\label{sec:threat}
In our threat model, we assume the client has purchased the trained DNN model infected with Trojans from a malicious party. Accordingly, we consider the following constraints on our defense strategy: (1)~The client has access to model weights but not the training data. (2)~The client has access to clean test data but they are unlabeled. (3)~The client is not aware whether or not the model is infected with Trojans. (4)~No prior knowledge is available about possible Trojan trigger shapes and/or patterns. 

To construct the defense, we assume access to a small corpus of \textit{unlabeled} data\footnote{less than 1\% of the training set size across all of our evaluations}. This is a realistic assumption as access to small amounts of data is possible via online resources. For instance, publicly available repositories enable data generation through generative networks for Faces\footnote{\url{http://www.whichfaceisreal.com/index.php}}. We consider the most generic and challenging form of Trojan attacks in which the attacker can control the trigger size, shape, and content. 
\sys{} mitigation is made possible in such scenario by constructing the defense using benign unlabeled data.

\section{\sys{} Components}
\subsection{DCT extraction}\label{sec:DCT}
\begin{figure}[h]
    \centering
    \vspace{-0.3cm}
    \includegraphics[width=0.99\columnwidth]{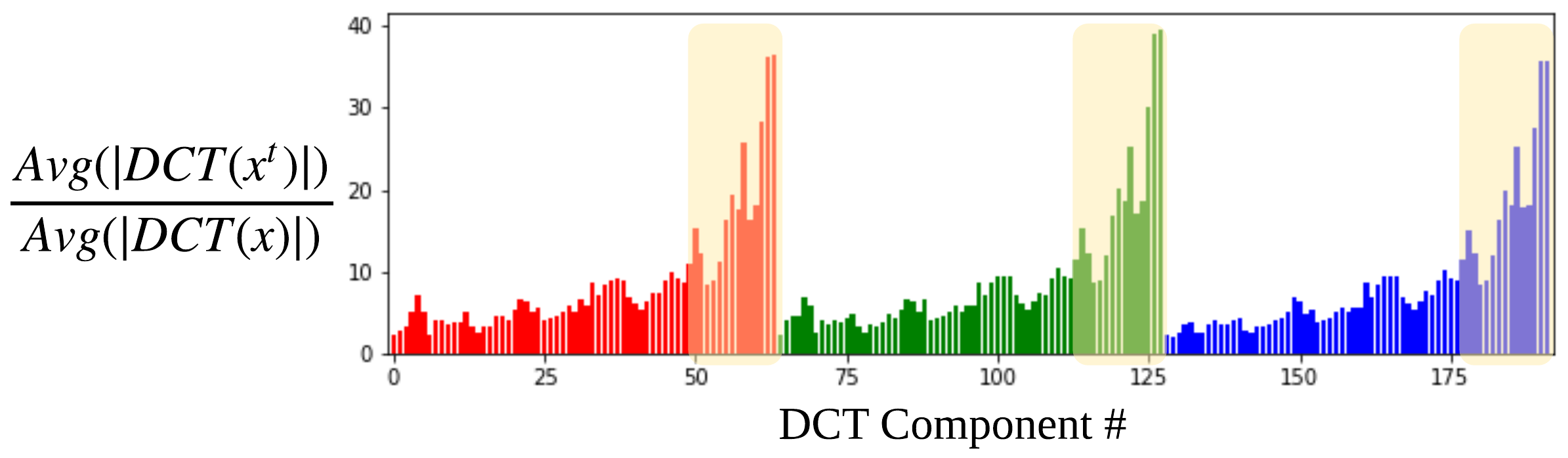}
    \vspace{-0.3cm}
    \caption{Average magnitude of DCT components for Trojan samples, normalized by benign data, shown in the three RGB channels. Trojans contain abnormally larger amounts of high-frequency components (highlighted regions).}
    \label{fig:dct_features}
    \vspace{-0.3cm}
\end{figure}

In natural images, most of the energy is contained in low frequencies. However, this property does not necessarily hold true for the Trojan triggers. Figure~\ref{fig:dct_features} shows the visualization of the frequency components for a Trojan sample, normalized by the magnitude of frequency components for benign data. Here, the magnitudes are averaged across $100,000$ image patches and the Trojan samples contain a watermark trigger generated by~\cite{liu2017trojaning}. As seen, Trojans have much larger components in the high-frequency domain compared to benign samples.

To perform frequency analysis, we divide each input image into non-overlapping patches\footnote{We use $P=4$ for small image benchmarks where input image dimensions are less than $32$ pixels. For larger input image sizes we use $P=8$.} of size $P\times P$. We transform each image patch to another patch of same size in the frequency domain using DCT.  
Eq.~(\ref{eq:dct}) encloses the formula used to compute the DCT transformation $F_{u,v}$ of a $P\times P$ patch.

\begin{equation}\label{eq:dct}
\resizebox{0.9\columnwidth}{!}{
    $F_{u,v} = C_{u,v}\sum\limits_{i=0}^{P-1}\sum\limits _{j=0}^{P-1} x_{i,j}\cos \left[{\frac {u~\pi }{P}}\left(i+{\frac {1}{2}}\right)\right]\cos \left[{\frac {v~\pi }{P}}\left(j+{\frac {1}{2}}\right)\right]$
}
\end{equation}

Here, $x_{i,j}$ is the input pixel located at the $(i,j)$ coordinate
and $C_{u,v}$ is a scalar constant that depends on the frequency coordinates. 
The extracted $2D$ DCT components $F_{u,v}$ are then sorted in decreasing order in terms of the information they carry following a zigzag pattern~\cite{salomon2004data}.

We represent the DCT transform as a group convolution with kernel size $P$ and $c_{in}$ groups where $c_{in}=3$ and $1$ for RGB and gray-scale images, respectively. The kernel weights of the convolution layer are initialized with the DCT basis coefficients which are pre-computed based on Eq.~(\ref{eq:dct}). The stride of the convolution is set to $P$ to account for image patching.
Such representation allows for an efficient implementation of the DCT Analyzer, which can be easily integrated into the architecture of the victim model as a pre-processing layer.


\subsection{Sparse Recovery}\label{sec:dict}

Sparse coding is referred to learning methods where the goal is to efficiently represent the data using sets of over-complete bases. Given a matrix of ($n$) data observations $X\in \mathbb{R}^{l\times n}$, sparse coding extracts a dictionary of normalized basis vectors $D\in \mathbb{R}^{l\times m}$ and the sparse representation matrix $V\in \mathbb{R}^{m\times n}$. Formally, the sparse coding objective can be written as:
\begin{equation}
    \underset{D,V}{min}~f_D(X) = \underset{D,V}{min}\|X - D.V\|_2 + \gamma\|{V}\|_0
\end{equation}
where $\gamma$ is a regularization coefficient that promotes sparsity in the coded representation $V$. Dictionary learning algorithms provide solutions to the above optimization problem by finding a dictionary $D$ that minimizes $\mathbb{E}_{x\sim \mathcal{X}}f_D(x)$,
where $\mathcal{X}$ is the distribution over the inputs. \sys{} extracts $D$ by performing dictionary learning over legitimate (benign) data. The out-of-distribution Trojan samples are thus expected to show a high reconstruction error, whereas benign samples will be accurately reconstructed with small error. 

\begin{SCfigure}[5][h]
    \centering
    \includegraphics[width=0.45\columnwidth]{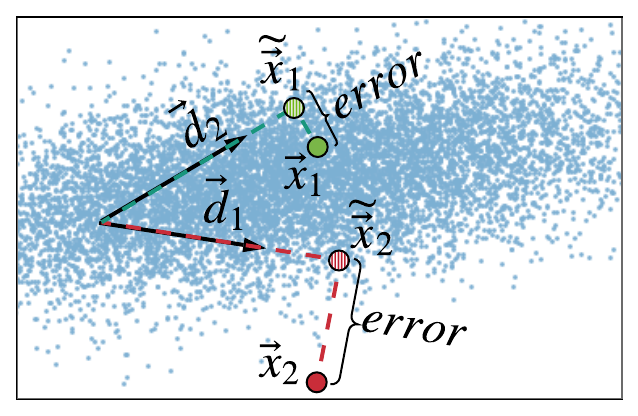}
    \caption{Illustration of sparse reconstruction for regular data (green circle) and out-of-distribution samples (red circle).}
    \label{fig:dict_outlier}
\end{SCfigure}

Figure~\ref{fig:dict_outlier} illustrates this behavior in an example $2D$ space. The light-blue dots represent the distribution of benign samples; the two solid arrows $\vec d_1$ and $\vec d_2$ are the dictionary atoms and only one of them is used for sparse reconstruction $\widetilde{\vec x}$. 
As seen, the magnitude of the reconstruction error on the outlier sample $\vec x_2$ is larger than that of regular data $\vec x_1$, i.e., $\|\vec x_2-\widetilde{\vec x}_2\|_F>>\|\vec x_1-\widetilde{\vec x}_1\|_F$ where $\|\cdot\|_F$ is the Frobenius norm.

While the above simple illustration shows the effectiveness of dictionary learning in $2$ dimensions, a similar behavior is observed when generalizing sparse coding to higher dimensions.
For a dictionary trained on $n$ samples $x\sim \mathcal X$, there exist theoretical bounds on the generalization error for unseen samples drawn from the same distribution $\mathcal{X}$. Let us denote the average reconstruction error over the set of $n$ observed samples by $E_o$. The generalization error of the dictionary $E_D(\cdot)$ on unseen samples $x_u\sim X$ is bounded by $E_D(x_u)\leq E_o + \delta$.
Vainsencher et al.~\cite{vainsencher2011sample} prove that the generalization error $\delta$ for a $\lambda$-sparse representation  is $\mathcal{O}(\sqrt{ml~\mathrm{ln}(n\lambda)/n})$ under some orthogonality assumptions for the dictionary. 
\sys{} dictionaries minimize reconstruction error on benign samples. We therefore carefully tune 
the dictionary size $m$ and sparsity level $\lambda$ to ensure a low reconstruction error on the data at hand ($E_o$) as well as a low error bound $\delta$.

\vspace{-0.3cm}
\vspace{-0.3cm}

\noindent{\textbf{$\blacktriangleright$ Data.}} We apply sparse recovery on two data subsets extracted from a small corpus of randomly selected benign samples. 
\begin{enumerate}[leftmargin=*]
    \item At the input of the neural network (Section~\ref{sec:DCT}), each column of matrix $X$ is the DCT of a single patch in the input image. For instance, for an $8\times 8$, DCT window, the dimensionality would be $l=3\times64$ ($64$ DCT coefficients per RGB channel). 
    \item At the latent space, each column of $X$ represents a flattened feature-map with reduced dimensionality.
\end{enumerate}

\noindent{\textbf{$\blacktriangleright$ Dictionary Learning.}} 
We use an adaptive sampling distribution based on the reconstruction error of $X$, dubbed Column Selection-based Sparse Decomposition (CSSD)~\cite{mirhoseini2017rankmap} for learning the dicrionaries. This algorithm initializes $D$ by a small random subset of $X$ and then iteratively adds columns to $D$; the probability of a data sample being appended at each step is proportional to its reconstruction error with the current column set. Formally, the probability of the $i$-th sample $x_i$ being selected at the $(t+1)$-th iteration is given by: 
\begin{equation}\label{eq:adaptive_sampling}
    p(i)\propto\frac{\|D_tD_t^+x_i-x_i\|_2}{\|x_i\|_2}
\end{equation}
where $D_t$ corresponds to the columns of the dictionary selected up to the $t$-th iteration and $D^+_t=(D_t^TD_t)^{-1}D_t^T$ is the pseudo inverse of $D_t$. 
The intuition behind Eq.~(\ref{eq:adaptive_sampling}) is to give a higher chance of selection to those elements of $X$ with higher reconstruction errors. This approach allows us to maximize the amount of embedded information from the data distribution inside $D$. 
While more sophisticated algorithms can be used~\cite{engan1999method,aharon2006k,aharon2008sparse}, our empirical evaluations show that CSSD can sufficiently express the data distribution with minimal generalization error. 

\noindent{\textbf{$\blacktriangleright$ Reconstruction Algorithm.}} We use Orthogonal Matching Pursuit (OMP)~\cite{davis1994adaptive} for sparse recovery as summarized in Algorithm~\ref{alg:OMP}. OMP iteratively finds non-zero elements to construct the sparse representation $\vec x$.
The added non-zero element at each iteration is chosen such that it minimizes the $L_2$ norm of the remaining residual error $\|\vec r_i-D^*\cdot\vec v\|_2$ which can be solved using Least-square (LS) optimization. The subset of dictionary columns ($D^*$) that contribute to the sparse recovery is also expanded over iterations. After $\lambda$ iterations, the reconstruction is returned as $\widetilde{\vec x}=D^*\cdot\vec v$.


\vspace{-0.2cm}
\setlength{\textfloatsep}{0pt}
\begin{algorithm}[h]
        \caption{OMP algorithm}\label{alg:OMP}
\begin{flushleft}
        \textbf{Inputs:}  Dictionary $D \in \mathbb{R}^{l\times m}$, input sample $\vec x\in\mathbb{R}^l$, number of non-zero coefficients for sparse recovery ($\lambda$).\\
        \textbf{Output:} reconstruction $\widetilde{\vec x}\in\mathbb{R}^l$.
\end{flushleft}
    \begin{algorithmic}[1]
        \State  $\vec r_0 \gets \vec x$ 
        \Comment{residual error: $\vec r_0\in\mathbb{R}^{l}$}
        \State  $D^* \gets \emptyset$ \Comment{empty dictionary subset}
        \For  {$i = 0, \dots, (\lambda-1)$}
        \State $\vec p = |D\cdot\vec{r}_{i}|$ \Comment{projection vector: $\vec p\in\mathbb{R}^{m}$}
        \State $j=\mathrm{argmax} \ \vec p$
            \State $D^* \gets D^* \cup  D\scriptstyle{[:,j]}$ \label{alg::OMP_bestcol} \Comment{update dictionary subset}
            \State  ${\vec v} \gets \mathrm{argmin}\  \|r_{i} - D^*\cdot \vec v\|_2$
            \State $\vec r_{i+1} \gets \vec r_{i}-D^* \cdot \vec v$\Comment{update residual error} \label{alg::OMP_res}
        \EndFor 
        \State \textbf{return} $D^*\cdot \vec v$ 
    \end{algorithmic}
\end{algorithm}

\vspace{-0.3cm}
\noindent{\textbf{$\blacktriangleright$ Distribution Learning with Few Samples.}} An ``over-complete'' dictionary is necessary to ensure representation sparsity~\cite{mirhoseini2017rankmap} and effective separation of outlier and benign samples. The term over-complete is used when the number of columns in the dictionary is higher than the data dimensionality ($m>>l$).  In real-world DNN applications, however, the number of data samples ($m$) is often small while the feature-map dimensionality ($l$) is large. To tackle this, we apply Singular Value Decomposition on the high-dimensional feature-maps to reduce $l$. Inverse SVD can then be applied on the reconstructed output to recover the original dimensionality. We choose the SVD rank such that more than $90\%$ of the original energy is preserved.

\vspace{-0.3cm}
\subsection{Outlier Detection}\label{sec:outlier}

As discussed in Section~\ref{sec:dict}, 
we leverage the disparity between the reconstruction error of benign and Trojan samples after undergoing sparse recovery to detect Trojans. Towards this goal, we first extract the statistical properties of the reconstruction error across benign samples. 
The out-of-distribution samples, i.e., outliers, are then marked as Torjan.
In order to model out of distribution samples, we utilize a multivariate extension of Chebyshev's inequality ~\cite{stellato2017multivariate}. Consider a random variable $\mathcal{X}\in\mathbb{R}^{1\times d}$ and let $\{\vec x_i\}_{i=1}^N$ denote a set of observed samples drawn from $\mathcal{P_\mathcal{X}}$. 
Based on the $N$ observations, we calculate the empirical mean $\vec\mu$ and the covariance $\Sigma$ as follows:
\begin{align}\label{eq:mean_cov}
    \vec \mu = \frac{1}{N}\sum_{i=1}^N \vec x_i\ , \quad
    \Sigma = \frac{1}{N-1}\sum_{i=1}^N (\vec x_i-\vec\mu)(\vec x_i-\vec\mu)^T
\end{align}

The Chebyshev's inequality provides an upper bound on the probability of samples lying outside ellipsoids of the form
$(\vec x-\vec \mu )\Sigma^{-1}(\vec x-\vec \mu)^T = \epsilon^2$. Let us denote the \textit{distance} of each sample from the distribution by:
\begin{equation}\label{eq:dist}
    dist(\vec x)=(\vec x-\vec \mu)\Sigma^{-1}(\vec x-\vec \mu)^T
\end{equation}
The Chebyshev's inequality can then be formally written as:
\begin{equation}\label{eq:prob}
\mathcal{P} (dist \geq \epsilon^2)\leq min\bigg\{ 1, \frac{d(N^2-1+N\epsilon^2)}{N^2\epsilon^2} \bigg\}
\end{equation}
The above inequality implies that one can categorize samples satisfying large enough values of $\epsilon$ as out-of-distribution, i.e., outlier. Based on this intuition, we measure the empirical mean and covariance in Eq.~(\ref{eq:mean_cov}) on a held-out dataset of benign samples and use the Chebyshev's inequality to characterize Trojaned data that do not belong to the benign probability distribution. The right-hand side of Eq.~(\ref{eq:prob}) provides the probability of a benign sample being categorized as outlier or Trojan. For large-enough values of $N$ ($N\rightarrow\infty$), this probability tends to $min\big\{1, \frac{d}{\epsilon^2}\big\}$. 

\begin{figure}[h]
    \centering
    \vspace{-0.2cm}
    \includegraphics[width=0.78\columnwidth]{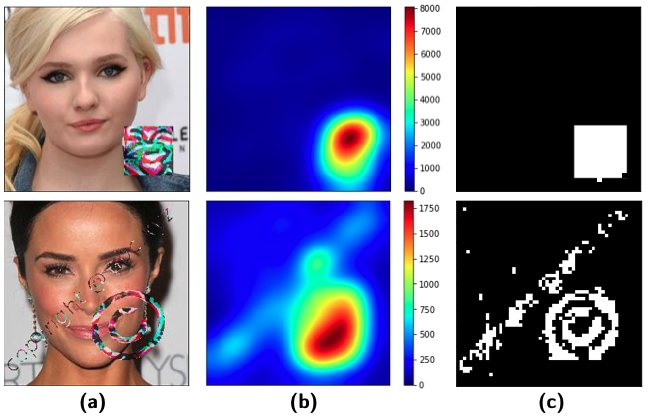}
    \vspace{-0.3cm}
    \caption{(a) Example Trojan data with watermark and square triggers~\cite{liu2017trojaning}, (b) reconstruction error heatmap, and (c) output mask from the outlier detection module.
    }\label{fig:mask}
    \vspace{-0.3cm}
\end{figure}

Figure~\ref{fig:mask}-a, b illustrates example Trojan data together with the corresponding reconstruction error heat maps. As seen, the Trojan trigger patterns have relatively larger reconstruction error compared to the rest of the image. 
Figure~\ref{fig:mask}-c visualizes the output of the outlier detection. Here, we generate a binary mask 
where the values of $0$ and $1$ correspond to in-distribution and outlier labels, respectively. As seen, parts of the input image that are covered with the Trojan trigger are correctly distinguished from benign regions. 

\noindent\textbf{$\blacktriangleright$ Tuning the parameter $\epsilon$.} 
We provide a systematic way to tune the parameter $\epsilon$ for outlier detection, based on the user-defined constraints on Trojan defense performance. 
An incoming sample $I\in \mathbb{R}^{d\times K\times K}$ is labeled as Trojan if at least one of its enclosing components $I_k\in\mathbb{R}^{d}$ 
is categorized as an outlier based on Eq.~(\ref{eq:prob}). The probability of an image being categorized as Trojan is therefore:
\begin{equation}\label{eq:fpr}
    \mathcal{P}_I(Trojan) = 1 - \prod_{k=1}^{K\times K} \mathcal{P}_{I_k}(Benign)
\end{equation}

When examining the outlier detection scheme on benign samples, the left-hand side of Eq.~(\ref{eq:fpr}) is equivalent to the False Positive Rate (FPR), i.e., the probability of a benign image being mistaken for a Trojan. Eq.~(\ref{eq:prob}) provides that for benign samples $\mathcal{P}_{I_k}(Benign|I_k\in Benign) \geq 1 - \frac{d}{\epsilon^2}$. The FPR is thus upper-bounded by:
\begin{equation}
    FPR = \mathcal{P}_I(Trojan|I\in Benign) \leq 1 - \Big(1-\frac{d}{\epsilon^2}\Big)^{K\times K}
\end{equation}
We can therefore determine the parameter $\epsilon$ based on the desired application-specific FPR denoted by $FPR_{target}$:
\begin{align}
    \underset{\epsilon}{\sup}~FPR = 1 - \Big(1-\frac{d}{\epsilon^2}\Big)^{K\times K} &\leq FPR_{target} \\
    \Rightarrow \frac{d}{\epsilon^2}\leq 1 - \sqrt[K\times K]{1-FPR_{target}}
\end{align}
where $\frac{d}{\epsilon^2}$ is the per-patch FPR, i.e., $\mathcal{P}_{I_k}(Trojan|I_k \in Benign)$. 

\begin{figure}[h]
    \centering
    \vspace{-0.2cm}
    \includegraphics[width=0.7\columnwidth]{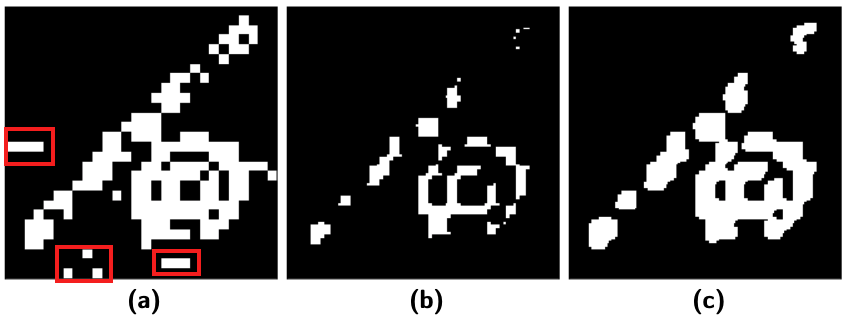}
    \vspace{-0.3cm}
    \caption{(a) Binary Trojan mask with the red rectangles indicating False alarms. (b) Output mask obtained after applying $2D$ binary erosion. (c) Output mask after restoring the high-concentration Trojan regions with $2D$ binary dilation. }\label{fig:morphology}
    \vspace{-0.5cm}
\end{figure}

\vspace{0.2cm}
\noindent\textbf{$\blacktriangleright$ Reducing FPR with Morphological Transforms.} As seen in Figure~\ref{fig:mask}, certain benign elements in the samples might be marked as Trojan, thus increasing the FPR. To reduce such patterns, we utilize two operations from morphological image processing, namely, erosion and dilation, implemented as convolution layers. 
Erosion emphasizes contiguous regions in the input mask and removes small, disjoint regions. Once erosion is applied, binary dilation restores high-density non-zero regions in the original input mask. 
Figure~\ref{fig:morphology}-a demonstrates the obtained binary mask from the outlier detection where the benign regions mistaken for being Trojan are marked with red boxes around them. 
Figure~\ref{fig:morphology}-b shows how erosion successfully removes the false alarms and Figure~\ref{fig:morphology}-c demonstrates how dilation restores the original shape of the binary mask in Trojan regions. 

\vspace{-0.2cm}
\subsection{Decision Aggregation}\label{sec:fusion}
\begin{figure}[h]
    \centering
    \vspace{-0.3cm}
    \includegraphics[width=0.99\columnwidth]{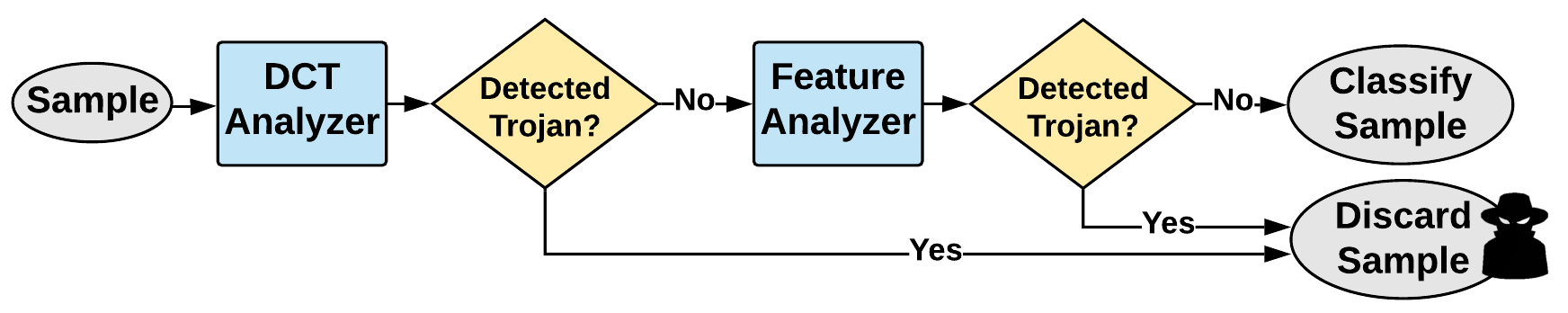}
    \vspace{-0.3cm}
    \caption{Decision flowchart for Trojan detection in \sys{}.}
    \vspace{-0.2cm}
    \label{fig:detect_chart}
\end{figure}

Figure~\ref{fig:detect_chart} illustrates the decision flowchart for \sys{} Trojan detection. As shown, a successful Trojan attack needs to satisfy two conditions: (1)~both the DCT and feature analyzers mistakenly mark the sample as benign, and (2)~the victim model classifies the sample in the target Trojan class. For each Trojan sample $x_i^t$, the attack success $S_i$ is computed as:
\begin{equation}\label{eq:trojan}
    S_i = (1-d_{DA}(x_i^t))(1-d_{FA}(x_i^t))(\mathcal{M}(x_i^t)\!==\!c^t)
\end{equation}
where $d_{DA}(\cdot)$ and $d_{FA}(\cdot)$ denote the decision of the DCT and feature analyzer modules, respectively, with the value of $1$ meaning the Trojan has been detected. Here, $\mathcal{M}(\cdot)$ represents the classification decision made by the victim model and $c_t$ is the Trojan attack target class. The overall attack success rate (ASR) is the expectation of $S$ over Trojan samples ($x^t\sim \mathcal{X}^t$). Since the three terms in Eq.~(\ref{eq:trojan}) are independent, we can write ASR as:
\begin{equation}\label{eq:asr_exp}
\resizebox{0.9\columnwidth}{!}{
    $ASR = \mathbb{E}_{\mathcal{X}^t}(1-d_{DA})\times \mathbb{E}_{\mathcal{X}^t}(1-d_{FA})\times \mathbb{E}_{\mathcal{X}^t}(\mathcal{M}(x_i^t)\!==\!c^t)$
}
\end{equation}

The first and second terms in the equation above are quantified using the True Positive detection rate (TPR). In this context, TPR measures the ratio of Trojan samples that are correctly identified by the defense. Let us denote the TPR for the DCT and feature analyziers with $TPR_{DA}$ and $TPR_{FA}$, respectively. Eq.~(\ref{eq:asr_exp}) can then be equivalently written as:
\begin{equation}\label{eq:asr}
    ASR = (1-TPR_{DA})(1-TPR_{FA})
    \times \frac{1}{N}\sum_{i=1}^N
    (\mathcal{M}(x_{i}^t)\!==\!c^t)
\end{equation}

Similarly, the classification accuracy on benign samples $ACC-C$ can be written in terms of the FPR of the DCT and feature analyzers:
\begin{equation}
    ACC-C = (1-FPR_{DA})(1-FPR_{FA})\times \frac{1}{N}\sum_{i=1}^N(\mathcal{M}(x_{i})\!==\!c_i)
\end{equation}
where $c_i$ denotes the correct class for the $i-$th sample.

\vspace{-0.1cm}
\section{\sys{} Hardware}\label{sec:hw}

In the following, we delineate the hardware architecture of \sys{} components that enable a high throughput and low energy execution. 

\begin{figure}[h]
    \centering
    \vspace{-0.4cm}
    \includegraphics[width=0.9\columnwidth]{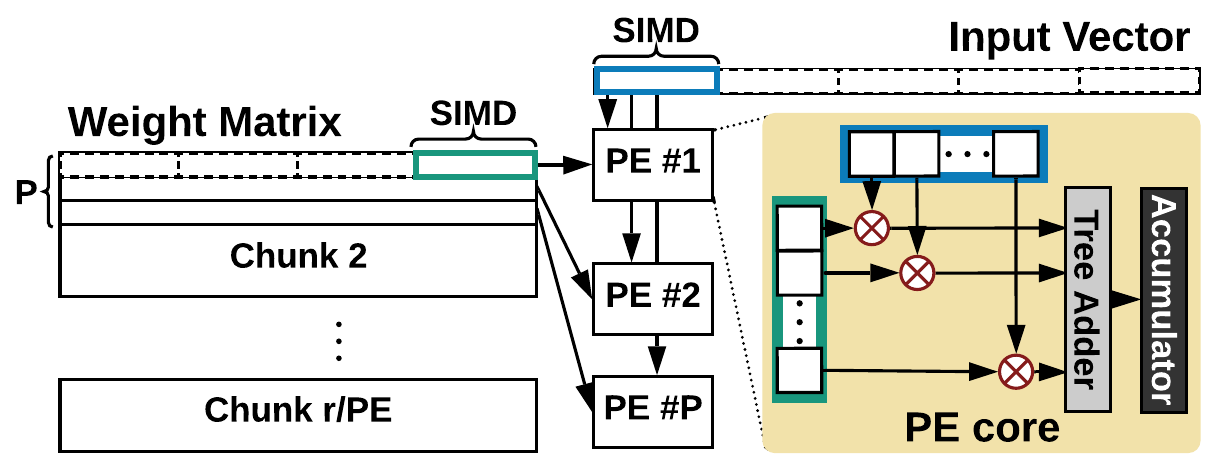}
    \vspace{-0.2cm}
    \caption{Schematic representation of \sys{} \texttt{MVM} core with its internal parallelization levels.}\label{fig:matmul}
\end{figure}

\noindent\textbf{$\blacktriangleright$ Matrix-Vector Multiplication Core.} Many of the fundamental operations performed in \sys{} include matrix-vector multiplication (\texttt{MVM}). In particular, the outlier detection module requires two \texttt{MVM}s to calculate the distance function shown in Eq.~(\ref{eq:dist}). Additionally, the dimensionality reduction and restoring components in the feature analyzer are realized using \texttt{MVM}s with weight matrices $W\in\mathbb{R}^{l\times r}$ and $W\in\mathbb{R}^{r\times l}$, respectively, where $l$ is the dimensionality of the input and $r$ is the SVD rank.
We devise an FPGA core for \texttt{MVM} and vector addition, realized using DSP blocks with 
Multiplication Accumulation (MAC) functionality~\cite{samraghencodeep,hussain2019fastwave}. 
Figure~\ref{fig:matmul} presents the high-level schematic of \sys{} vector-matrix multiplication. 

We provide two levels of parallelism in our design controlled by parameters \texttt{P} and \texttt{SIMD} in figure~\eqref{fig:matmul}.
This approach allows our design to achieve maximum resource utilization and throughput on various FPGA platforms.
The weight matrix is divided into subsets of length \texttt{P} and fed into parallel processing elements (\texttt{PE}s). These subsets are read from DRAM using a Ping-Pong weight buffer to overlap memory reads with \texttt{PE} computations.
At each cycle, \texttt{PEs} perform partial dot-product on the fetched weight and input partitions of length $SIMD$; the same input partition is shared across all \texttt{PE}s. We devise a tree-based reduction module and an accumulator to enable summation of partial dot-product outputs.

\noindent\textbf{$\blacktriangleright$ Sparse Recovery Core.}
The sparse recovery module performs OMP to reconstruct input signals.
We provide a reconfigurable and scalable OMP core on FPGA to accelerate sparse recovery. OMP relies on sequential execution of three steps: (1)~The dictionary column with the maximum dot-product with the current residual vector is selected. (2)~An LS optimization step generates the sparse representation of the current residual vector with the columns of the dictionary selected so far. (3)~The residual is updated based on the new sparse representation and the selected dictionary columns.  

We utilize \sys{} \texttt{MVM} core to implement the first step above. For the second step, we implement the LS optimization using a $QR$ factorization of the dictionary matrix. We leverage the modified Gram Schmidt (MGS) method~\cite{golub2012matrix} to perform the factorization. Since the dictionary matrix expands by one column each iteration, it is not necessary to recompute the $Q$ and $R$ matrices very time. Instead, we iteratively form the $Q$ and $R$ matrices as outlined in Algorithm~\ref{alg:OMP_QR}. Using the acquired new column for the $Q$ matrix, the residual update step takes the following form:
\begin{equation}\label{eq:QR_res}
    \vec r_{i+1} \gets \vec r_{i} - Q_i(Q_i)^T \vec r_{i}
\end{equation}
Due to the low memory footprint of \sys{} components, we store all required data in the available on-chip Block RAMs. By eliminating the overhead of external memory access, \sys{} enjoys a low latency and high power efficiency.

\vspace{-0.2cm}
\setlength{\textfloatsep}{0pt}
\begin{algorithm}[h]
\caption{QR factorization with MGS}\label{alg:OMP_QR}
\begin{flushleft}
        \textbf{Inputs:}  New dictionary column $D^*_{i}$, $Q_{i-1}$, $R_{i-1}$.\\
        \textbf{Output:} $Q_i$, $R_i$.
\end{flushleft}
\begin{algorithmic}[1]
    \State  $R_i \gets \begin{bmatrix} R_{i-1} & 0 \\ 0 & 0 \end{bmatrix} , ~\epsilon_i \gets D^*_{i}$
    \For  {$j = 1,\cdots,(i-1)$}
        \State $R_i \scriptstyle[j,i] \textstyle \gets (Q_{i-1}\scriptstyle[j]\textstyle)^T\epsilon_i$ 
        \State  $\epsilon_i \gets \epsilon_i - R_i\scriptstyle[j,i]\textstyle Q_{i-1}\scriptstyle[j]$ 
    \EndFor 
    \State $R_i\scriptstyle[i,i] \textstyle= \|\epsilon_i\|_2$
    \State $Q_i = Q_{i-1} \cup \epsilon_i/R_i\scriptstyle[i,i]$
\end{algorithmic}
\end{algorithm}
\vspace{-0.5cm}




\section{Experiments}\label{sec:experiments}
We evaluate \sys{} on three visual classification datasets of varying size and complexity, namely, MNIST~\cite{MNIST} for handwritten digits, GTSRB~\cite{Stallkamp2012} for road signs, and VGGFace~\cite{parkhi2015deep} for face data. The number of classes for each dataset is $10$, $43$, and $2622$, respectively. We corroborate \sys{} effectiveness against variations of two available state-of-the-art Neural Trojan attacks. In what follows, we provide detailed performance analysis and comparisons with prior art. We further demonstrate \sys{} accelerated execution on embedded hardware.


\vspace{-0.2cm}
\subsection{Attack Configuration}
Throughout the experiments, we consider input-agnostic Trojans where adding the trigger to any image causes misclassification to the attack target class. 
Table~\ref{tab:datasets} summarizes the evaluated benchmarks along with their corresponding Trojan attacks and triggers. 

\noindent\textbf{$\blacktriangleright$ BadNets.} We implement the BadNets~\cite{gu2017badnets} attack with various triggers as an example of a realistic physical attacks. The injected Trojans include a white square and a Firefox logo placed at the bottom right corner of the input image. We embed the backdoor by injecting $\sim 10\%$ poisoned data samples during training.

\noindent\textbf{$\blacktriangleright$ TrojanNN.}
We evaluate \sys{} against TrojanNN~\cite{liu2017trojaning} as a digital attack with complex triggers. The attack is implemented using the open-source models shared by TrojanNN authors\footnote{\url{https://github.com/PurduePAML/TrojanNN}}. We perform experiments with two variants of TrojanNN triggers, namely, square and watermark, crafted for the VGGFace dataset.

\begin{table}[h]
\vspace{-0.2cm}
\caption{Evaluated datasets and attack algorithms.}\label{tab:datasets}
\vspace{-0.3cm}
\resizebox{0.95\columnwidth}{!}{
\begin{tabular}{ccccc}
\hline
\textbf{Dataset} & \textbf{Input Size} & \textbf{Architecture} & \textbf{Attack} & \textbf{Trigger}                                           \\ \hline
\textbf{MNIST}   & 1x28x28             & 2CONV, 2MP, 2FC       & BadNets         & square                                                     \\ \hline
\textbf{GTSRB}   & 3x32x32             & 6CONV, 3MP, 2FC       & BadNets         & \begin{tabular}[c]{@{}c@{}}square\\ Firefox\end{tabular}   \\ \hline
\textbf{VGGFace} & 3x224x224           & 13CONV, 5MP, 3FC      & TrojanNN        & \begin{tabular}[c]{@{}c@{}}square\\ watermark\end{tabular} \\ \hline
\end{tabular}
}
\vspace{-0.6cm}
\end{table}

\subsection{Detection Performance}

We apply \sys{} Trojan mitigation at the input and latent space of infected DNNs. To create the defense, we separate 500, 430, and 2622 clean samples from MNIST, GTSRB, and VGGFace test sets, respectively. The aforementioned size for the benign dataset corresponds to $1\%$ of the training data size for MNIST and GTSRB and $0.1\%$  VGGFace training data. Such low data size requirements provide a competitive advantage for \sys{} defense in real-world scenarios. We summarize other defense parameters for our evaluated benchmarks in Table~\ref{tab:defense_parameters}. These parameters are selected to maintain a high classification accuracy over the benign data.

\begin{table}[h]
\vspace{-0.2cm}
\caption{Parameters of \sys{} modules for various datasets. $P$: DCT windows size, $l$: feature size for sparse recovery, $m:$ number of dictionary columns for sparse recovery, $\lambda$: sparsity parameter in sparse recovery, $\epsilon^2$: distance threshold for outlier detection.}\label{tab:defense_parameters}
\vspace{-0.2cm}
\resizebox{\columnwidth}{!}{
\begin{tabular}{cc|ccccc|cccc}
\hline
\multirow{2}{*}{\textbf{Dataset}} & \multirow{2}{*}{\textbf{Trigger}} & \multicolumn{5}{c|}{\textbf{Input Analyzer}}                                                                     & \multicolumn{4}{c}{\textbf{Feature Analyzer}}                     \\
&      & $P$     & $l$    & $m$                     & $\lambda$      & $\epsilon^2$   & $l$                      & $m$   & $\lambda$ & $\epsilon^2$ \\ \hline
\textbf{MNIST}                    & Square                            & 4                  & 48                   & 1000                  & 5                  & $5\times10^{-4}$                  & 279                    & 500                   & 80     & $2\times10^{-3}$   \\ \hline
\multirow{2}{*}{\textbf{GTSRB}}   & Square                            & \multirow{2}{*}{4} & \multirow{2}{*}{48}  & \multirow{2}{*}{1000} & \multirow{2}{*}{5} & \multirow{2}{*}{$5\times10^{-4}$} & \multirow{2}{*}{85}  & \multirow{2}{*}{420}  & 80     & $3\times10^{-3}$    \\
                                  & FireFox                           &                    &                      &                       &                    &                         &                        &                       & 50     & $1\times10^{-2}$     \\ \hline
\multirow{2}{*}{\textbf{VGGFace}} & Square                            & \multirow{2}{*}{8} & \multirow{2}{*}{192} & \multirow{2}{*}{1000} & \multirow{2}{*}{5} & $5\times10^{-4}$ & \multirow{2}{*}{520} & \multirow{2}{*}{2622} & 80     & $1\times10^{-4}$   \\
                                  & Watermark                         &                    &                      &                       &                    &  $8\times10^{-4}$                       &                        &                       & 80     & $1\times10^{-4}$   \\ \hline
\end{tabular}
}
\end{table}

\vspace{-0.2cm}
We evaluate \sys{} Trojan resiliency on physical and digital attacks in Table~\ref{tab:comparison}. Specifically, under ``Defended Model'', we evaluate the drop in clean data accuracy (ACC$\downarrow$), the attack success rate (ASR), and 
Trojan ground-truth label recovery (TGR). In addition to our results, we include prior art performance in terms of the above-mentioned criteria. On MNIST, \sys{} achieves 0\% ASR, with only 0.1\% drop in clean data accuracy, outperforming the prior art. For GTSRB, \sys{} achieves an ASR of $0\%$  and a lower drop of accuracy compared to all prior work, except for Deep Inspect, which suffers from a much higher ASR of $8.8\%$. 

On digital attacks, \sys{} achieves 0.0\% ASR with only 0.8\% and 2.0\% degradation of accuracy for square and watermark shapes. The watermark trigger covers a large area of the input image, obstructing the critical features. As such, while \sys{} detects the Trojan with high success, it shows a lower TGR compared to our other triggers. Note that Neural Cleanse and Deep Inspect perform DNN training on synthetic datasets achieved with model inversion~\cite{fredrikson2015model}. As a result, their post-defense accuracy is not directly comparable with \sys{}, which does not perform DNN retraining. We emphasize that while such retraining contributes to accuracy, it may not be feasible in real-world applications.


\begin{table}[ht]
\vspace{-0.2cm}
\caption{Evaluation of \sys{} on various physical and digital attacks. Comparisons with state-of-the-art prior works, i.e., Neural Cleanse(NC)~\cite{wang2019neural}, Deep Inspect (DI)~\cite{chen2019deepinspect}, Februus~\cite{doan2019februus}, and SentiNet~\cite{chou2018sentinet} are provided where applicable. }\label{tab:comparison}
\vspace{-0.3cm}
\resizebox{\columnwidth}{!}{
\begin{tabular}{ccccccccc}
\hline
\multirow{2}{*}{\textbf{Dataset}}                                                              & \multirow{2}{*}{\textbf{Trigger}} & \multirow{2}{*}{\textbf{Work}} & \multirow{2}{*}{\textbf{Retrain}} & \multicolumn{2}{c}{\textbf{Infected Model}} & \multicolumn{3}{c}{\textbf{Defended Model}} \\
&       &     &   & \textbf{ACC-C}        & \textbf{ASR}        & \textbf{ACC$\downarrow$}  & \textbf{ASR} & \textbf{TGR} \\ \hline
\multirow{3}{*}{\textbf{\begin{tabular}[c]{@{}c@{}}MNIST\\ (Physical\\ Attack)\end{tabular}}}  & \multirow{3}{*}{\begin{tabular}[c]{@{}c@{}}Square\\ $(4\times 4)$\end{tabular}}           & NC                             & yes                                  & 98.5                  & 99.9                & 0.8           & 0.6          & NA           \\
&                                   & DI                             & yes                                  & 98.8                  & 100.0               & 0.7           & 8.8          & NA           \\
&    & \bluecell{\sys}   &  \bluecell{no}                                   & \bluecell 99.3                  & \bluecell 100.0               & \bluecell\textbf{0.1}           & \bluecell\textbf{0.0}            & \bluecell\textbf{98.7}         \\ \hline
\multirow{6}{*}{\textbf{\begin{tabular}[c]{@{}c@{}}GTSRB\\ (Physical\\ Attack)\end{tabular}}}  & \multirow{4}{*}{\begin{tabular}[c]{@{}c@{}}Square\\ $(4\times 4)$\end{tabular}}  & NC                             & yes   & 96.5   & 97.4  & 3.6  & 0.1          & NA           \\
&    & DI   & yes   & 96.1  & 98.9                & \textbf{-1.0}         & 8.8          & NA           \\
&     & Februus    & yes$^*$                               & 96.8    & 100    & 1.2           & \textbf{0.0}         & \textbf{96.5}           \\
&      & \bluecell\sys   & \bluecell no                                   & \bluecell 96.5         & \bluecell 99.4      & \bluecell 0.0          & \bluecell\textbf{0.0}            & \bluecell 94.7         \\ 
\cline{2-9} 
&    \multirow{2}{*}{\begin{tabular}[c]{@{}c@{}}Firefox\\ $(6\times 6)$\end{tabular}} 
&   \bluecell &   \bluecell  &   \bluecell    &   \bluecell & \bluecell  &   \bluecell   &  \bluecell  \\
& & \multirow{-2}{*}{\bluecell\sys}   & \multirow{-2}{*}{\bluecell no}                                   & \multirow{-2}{*}{\bluecell 92.6}                  & \multirow{-2}{*}{\bluecell 99.8}                & \multirow{-2}{*}{\bluecell 0.4}           & \multirow{-2}{*}{\bluecell 1.7}         & \multirow{-2}{*}{\bluecell 83.5}         \\ 
\hline
\multirow{7}{*}{\textbf{\begin{tabular}[c]{@{}c@{}}VGGFACE\\ (Digital\\ Attack)\end{tabular}}} & \multirow{4}{*}{\begin{tabular}[c]{@{}c@{}}Square\\ $(59\times 59)$\end{tabular}}           & NC                             & yes                                  & 70.8                 & 99.9                & \textbf{-8.4}          & 3.7          & NA           \\
&                                   & DI                            & yes                                  & 70.8                  & 99.9                & 0.7           & 9.7          & NA           \\
 &                                   & SentiNet$^{\ddagger}$                       & no                                   & NA                    & 96.5                  & NA            & 0.8          & NA           \\
&    & \bluecell \sys            & \bluecell no                                   & \bluecell 74.9                  & \bluecell 93.52               & \bluecell 0.8           & \bluecell\textbf{0.0}          & \bluecell\textbf{70.1}         \\ \cline{2-9} 
& \multirow{3}{*}{Watermark}        & NC                             & yes                                  & 71.4                  & 97.60               & \textbf{-7.4}         & \textbf{0.0}          & NA           \\
&                                   & DI                             & yes                                  & 71.4                  & 97.60               & 0.5          & 8.9          & NA           \\
&                                   & \bluecell \sys            & \bluecell no                                   & \bluecell 74.9                  & \bluecell 58.6                & \bluecell 2.0           & \bluecell\textbf{0.0}          & \bluecell\textbf{41.38}        \\ \hline
\end{tabular}
}
\flushleft{\scriptsize $^*$ Februus performs GAN training. \qquad $^{\dagger}$ SentiNet only reports results on LFW~\cite{Huang2012a} dataset.}
\vspace{-0.7cm}
\end{table}

\begin{figure}[h]
    \centering
    \vspace{-0.3cm}
    \includegraphics[width=0.85\columnwidth]{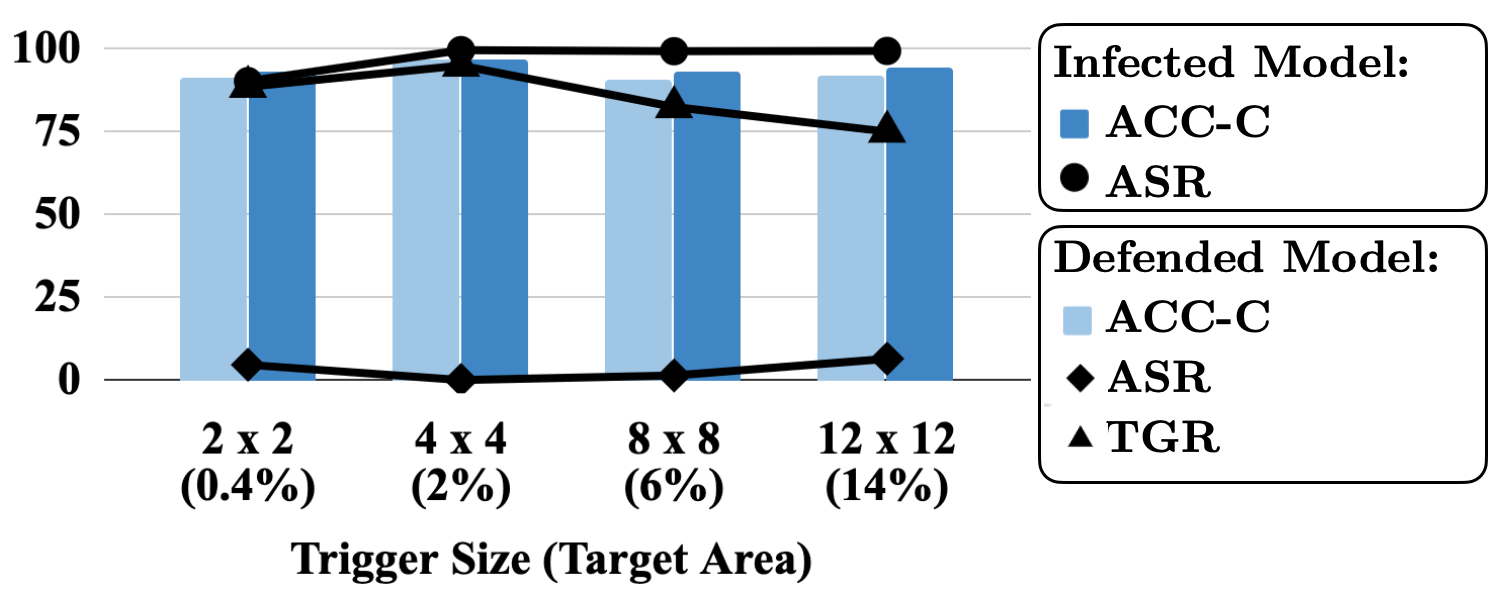}
    \vspace{-0.3cm}
    \caption{Analysis of \sys{} sensitivity to Trojan trigger size.}
    \label{fig:trigger_size}
    \vspace{-0.3cm}
\end{figure}

\vspace{0.5cm}
\noindent{\textbf{$\blacktriangleright$ Sensitivity to Trigger Size.}} 
We perform experiments on the GTSRB dataset with a square Trojan trigger and change the trigger size such that it covers between $\sim0.4\%$ to $\sim14\%$ of the input image area. The size range is chosen to ensure that the corresponding triggers are viable in real settings and provide a high ASR. We summarize the obtained results in Figure~\ref{fig:trigger_size}.
\sys{} significantly reduces the ASR while enabling recovery of ground-truth labels with a high accuracy across all trigger sizes. This is expected since \sys{} does not rely on the trigger size to construct the defense. 
For average sized Trojans, \sys{} successfully detects the existence of triggers and reduces the ASR to less than $1\%$. For larger trigger sizes, the TGR is relatively lower since the Trojan occludes the main objects in the image.

\noindent\textbf{$\blacktriangleright$ Offline Preprocessing Overhead.} 
The preparation of \sys{} defensive modules consists of the following steps:
\begin{itemize}
    \item DCT extraction and dictionary leaning on benign inputs.
    \item Computing $\vec\mu$ and $\Sigma$ in Eq.~(\ref{eq:mean_cov}) for input outlier detection.
    \item Computing SVD and dictionary learning at latent feature maps.
    \item Computing $\vec\mu$ and $\Sigma$ for latent outlier detection.
\end{itemize}
In practice, the above computation incurs negligible runtime compared to DNN training. We implement the above steps in PyTorch and measure the runtime on an NVIDIA TITAN Xp GPU. For our GTSRB benchmark, the above operations require $0.06$, $0.19$, $10.47$, and $0.1$ seconds, respectively. The defense construction time is therefore $\sim 11$ seconds which is $\sim 1.8\%$ of the time required to train the victim DNN on this benchmark. For the more complex VGGFace dataset, the above operations require $1.05$, $0.54$, $48.3$, and $1.2$ seconds, respectively, resulting in a total of $\sim51$ seconds for defense preparation.


\vspace{-0.2cm}
\subsection{Hardware performance}

We implement the proposed Trojan defense strategy on various hardware platforms and compare the performance of \sys{} components. The evaluated platforms include server-grade CPUs and GPUs, embedded CPUs and GPUs, and FPGA. We base our comparisons on performance-per-Watt defined as the throughput over the total power consumed by the system. This measure effectively encapsulates two major performance metrics for embedded applications.
Throughout this section, we will target our study on the GTSRB benchmark but similar trends are observed for other datasets.

\noindent\textbf{$\blacktriangleright$ Performance on General Purpose Hardware.} We provide an optimized software library for \sys{} defense components in Python. In order to benefit from highly optimized backend compilers for tensor operations on CPU and GPU, our codes are developed on top of the PyTorch deep learning library. Our provided software library can be readily instantiated within PyTorch API to enable simultaneous DNN execution and Trojan defense. We implement our defense pipeline on 
the \textit{Jetson TX2} embedded development board running in CPU-GPU and CPU-only modes. We further run the defense on a server-grade \textit{Intel Xeon E5} CPU and an \textit{NVIDIA TITAN Xp} GPU. 
The overall achieved defense throughput with a batch size of $1$ ranges from $11$ fps on the embedded CPU up to $28$ fps on the server GPU.

\begin{figure}[h]
\centering
\vspace{-0.3cm}
\includegraphics[width=0.99\columnwidth]{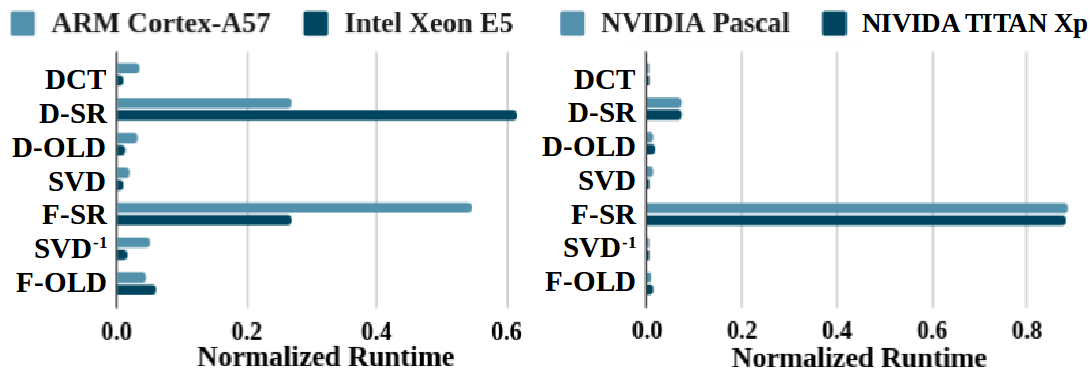}
\vspace{-0.2cm}
\caption{Latency breakdown of \sys{} components running on embedded and high-end CPUs (left) and GPUs (right).}\label{fig:runtime_CPU_GPU}
\vspace{-0.3cm}
\end{figure}

Figure~\ref{fig:runtime_CPU_GPU} illustrates the runtime breakdown for various components of \sys{} running on each platform. Here, the sparse recovery and outlier detection modules are abbreviated as \texttt{SR} and \texttt{OLD} and the prefixes \texttt{D-} and \texttt{F-} correspond to the DCT and feature analyzers, respectively. The experiments are performed using a batch size of $1$ to resemble real-world applications and runtimes are averaged across 100 runs. For each platform, we normalize the runtime of each component by the total defense execution time for one sample. As seen, the bulk of defense runtime belongs to the sparse recovery module. This is due to the inherently sequential nature of the OMP algorithm performed inside this module. CPU and GPU platforms are designed to excel in massively parallel operations while this does not hold for OMP. Such behavior further motivates us to design specialized hardware to accelerate the execution of \sys{} components on FPGA.

\noindent\textbf{$\blacktriangleright$ Performance on Customized Accelerator.} We implement \sys{} components on FPGA using the developed sparse recovery and \texttt{MVM} cores as the basic blocks. The design is developed in Vivado High-Level Synthesis and synthesized in Vivado Design Suite for the \textit{Xilinx UltraScale VCU108} board. Power consumption is estimated during synthesis with Vivado Design Suite. Finally, a comprehensive timing and resource utilization analysis is performed. To maximize throughput, we tuned the parallelism factors in the \texttt{MVM} modules to the highest value such that the design fits within the available resources. 

\begin{SCfigure}[5][h]
    \centering
    \includegraphics[width=0.5\columnwidth]{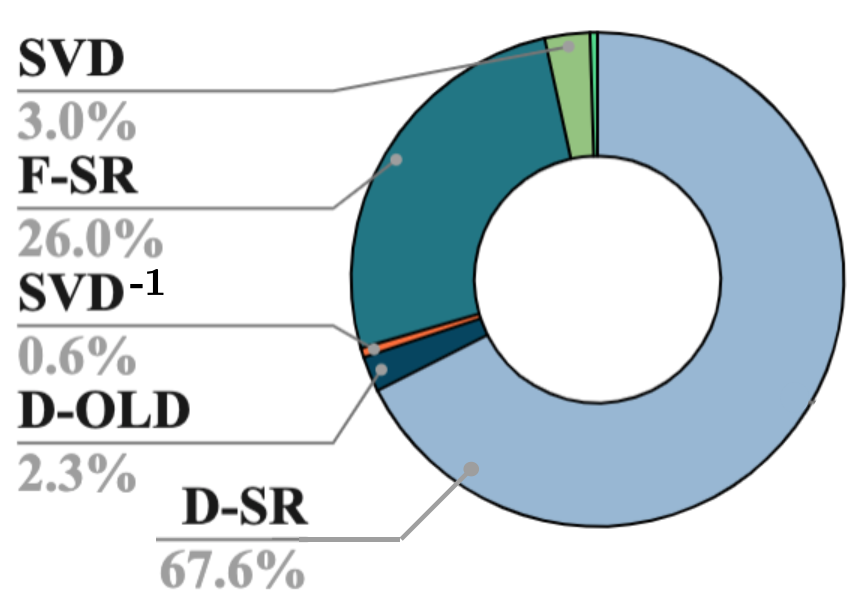}
    \caption{Cycle-count breakdown for execution of \sys{} components implemented on FPGA.}
    \label{fig:FPGA_pie}
\end{SCfigure}

Figure~\ref{fig:FPGA_pie} demonstrates the breakdown of execution cycles for \sys{} components. As seen, the sequential execution of the sparse recovery core accounts for the majority of computation cycles. Our FPGA-based sparse recovery core enjoys up to $10\times$ and $18\times$ faster execution, respectively, compared to their CPU and GPU counterparts. This is enabled by pipelined execution, fine-grained optimizations to data access patterns, and parallel computation.

\vspace{-0.3cm}
\vspace{-0.3cm}

\begin{figure}[h]
\centering
\includegraphics[width=0.99\columnwidth]{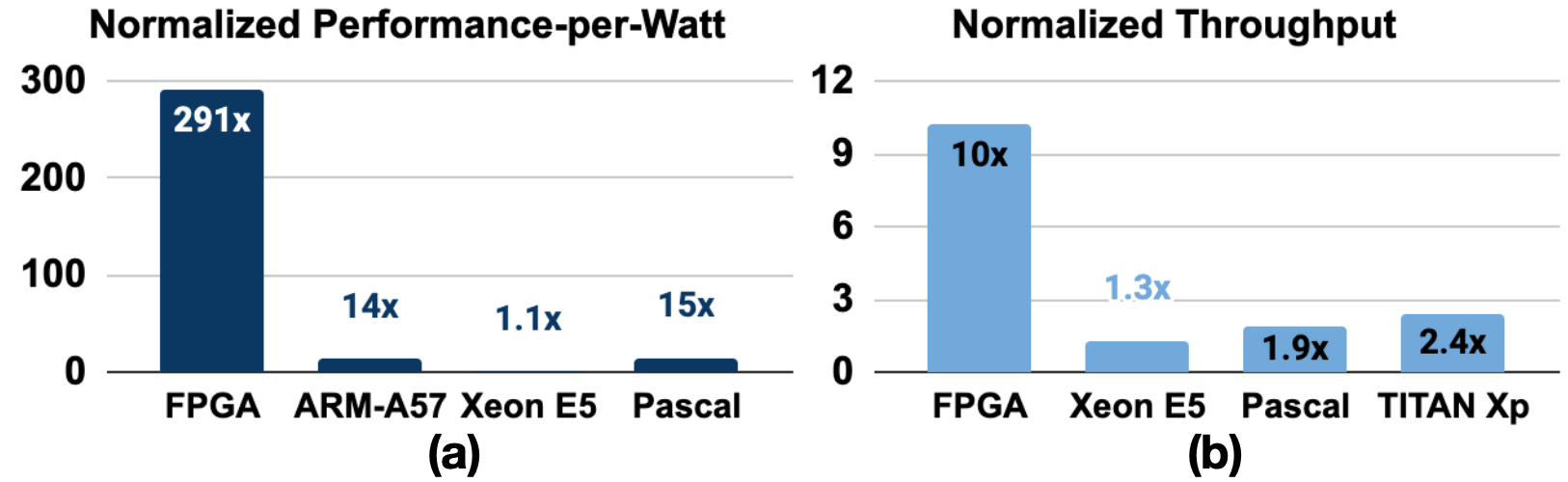}
\vspace{-0.3cm}
\caption{(a) Performance-per-Watt and (b) throughput across hardware platforms. Reported values for performance per-watt are normalized by \textit{TITAN Xp} and throughput values are normalized by \textit{ARM Cortex-A57}.}
\vspace{-0.5cm}
  \label{fig:throughputs}
\end{figure}

We compare the performance-per-Watt and throughput of \sys{} on different hardware platforms in Figure~\ref{fig:throughputs}. The performance-per-watt numbers are normalized by \textit{TITAN Xp} and the throughput numbers are normalized by \textit{ARM Cortex-A57}. As seen, the power-efficient implementation of \sys{} on FPGA not only enjoys a high throughput, but it also significantly increases performance-per-watt compared to commodity hardware. Note that due to the lightweight nature of \sys{} defense strategy, the server-grade GPU performs poorly in terms of performance-per-watt compared to other platforms due to under-utilization and excessive power consumption.

\section{Conclusion}\label{sec:conclusion}
This paper presents \sys{}, an end-to-end framework for online accelerated defense against Neural Trojans. The proposed defense strategy offers several intriguing properties: (1)~The defense construction is entirely unsupervised and sample efficient, i.e., it does not require any labeled data and is established using a small clean dataset. (2)~It is the first work to recover the original label of Trojan data without need for any fine-tuning or model training. (3)~\sys{} provides theoretical bounds on the false positive rate. (4)~The framework is devised based on algorithm/hardware co-design to enable accurate Trojan detection on resource-constrained embedded devices. We consider a challenging threat model where the attacker can use Trojan triggers with arbitrary shapes and patterns while no knowledge about the attack is available to the client. \sys{} light-weight defense and realistic threat model makes it an attractive candidate for practical deployment. Our extensive evaluations corroborate \sys{}'s competitive advantage in terms of attack resiliency and execution overhead. 

\vspace{-0.2cm}
\section{Acknowledgment}
This work was supported in part by ARO  (W911NF1910317).

\bibliographystyle{ACM-Reference-Format}
\bibliography{sample-bibliography} 

\end{document}